\pdfoutput=1

\documentclass[11pt]{article}

\usepackage[]{acl}

\usepackage{times}
\usepackage{latexsym}

\usepackage{graphicx}
\usepackage{color}
\usepackage{xspace}
\usepackage{amsmath,amsfonts}
\usepackage{algorithm}
\usepackage{algorithmic}
\usepackage{textcomp}
\usepackage{xcolor}
\usepackage{booktabs}
\usepackage{booktabs,xltabular}
\usepackage{subcaption}
\usepackage{listings}
\usepackage{adjustbox}
\usepackage[most]{tcolorbox}

\usepackage{url}

\newtcolorbox{promptBox}{
    colback=gray!10, 
    colframe=black!40, 
    fonttitle=\bfseries,
    coltitle=black,
    colbacktitle=gray!40
}

\lstdefinestyle{codestyle}{
    backgroundcolor=\color{white},   
    commentstyle=\color{green},
    keywordstyle=\color{blue},
    numberstyle=\tiny\color{black},
    stringstyle=\color{purple},
    basicstyle=\ttfamily\footnotesize,
    breakatwhitespace=false,         
    breaklines=true,                 
    captionpos=b,                    
    keepspaces=true,                 
    numbers=left,                    
    numbersep=5pt,                  
    showspaces=false,                
    showstringspaces=false,
    showtabs=false,                  
    tabsize=2
}
\usepackage[T1]{fontenc}

\usepackage[utf8]{inputenc}

\usepackage{microtype}

\newcommand{\TheName}{\textsc{Constructa}}

\title{\TheName{}: Automating Commercial Construction Schedules in
Fabrication Facilities with Large Language Models}

\author{
    \textbf{Yifan Zhang}\textsuperscript{1,2}\textsuperscript{\dag}\hspace{5pt}
    \textbf{Xue Yang}\textsuperscript{2}\vspace{5pt}\\
    \includegraphics[height=10pt]{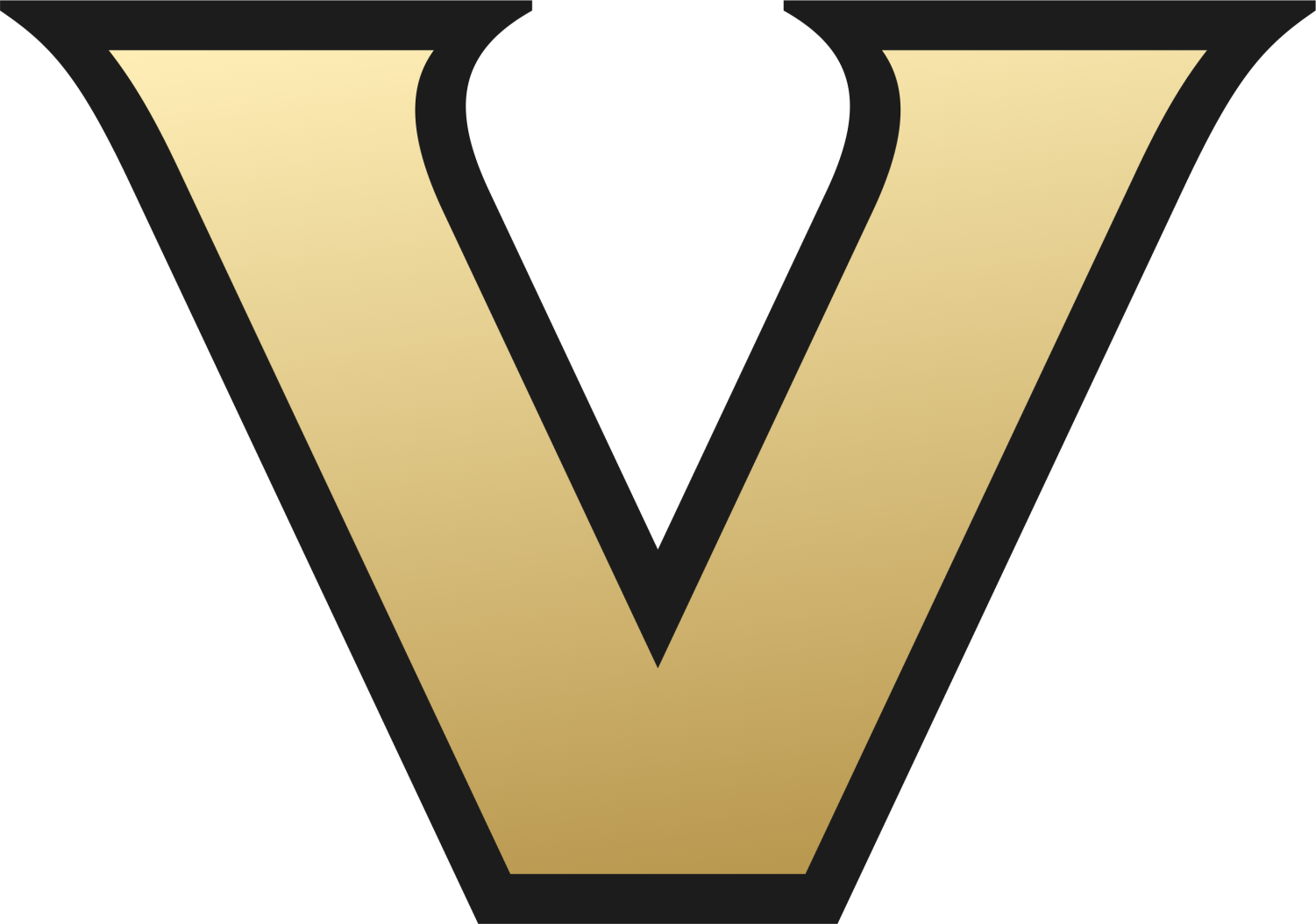} Vanderbilt University\textsuperscript{1}\hspace{5pt}
    \includegraphics[height=10pt]{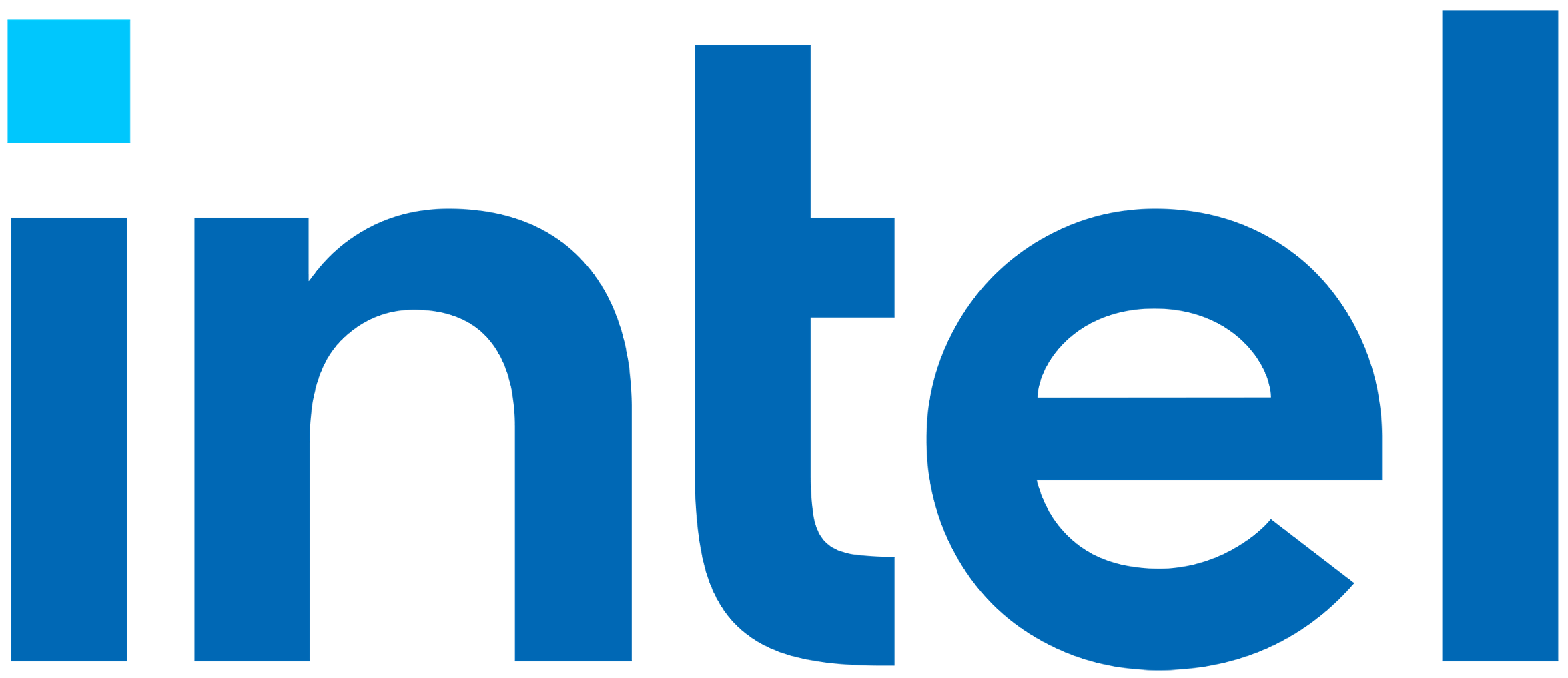} Intel Corporation\textsuperscript{2}\\
    \normalsize
    \texttt{\{yifan.zhang.2\}@vanderbilt.edu}\hspace{5pt} 
    \texttt{\{xue.yang\}@intel.com}
}

\begin{document}

\maketitle

\begingroup
\renewcommand\thefootnote{\dag}
\footnotetext{Work done during a GenAI research internship at Intel Incubation and Disruptive Innovation~(IDI) Group.}
\endgroup

\begin{abstract}

Automating planning with LLMs presents transformative opportunities for traditional industries, yet remains underexplored. In commercial construction, the complexity of automated scheduling often requires manual intervention to ensure precision. We propose \TheName{}, a novel framework leveraging LLMs to optimize construction schedules in complex projects like semiconductor fabrication. \TheName{} addresses key challenges by: (1) integrating construction-specific knowledge through static RAG; (2) employing context-sampling techniques inspired by architectural expertise to provide relevant input; and (3) deploying Construction DPO to align schedules with expert preferences using RLHF. Experiments on proprietary data demonstrate performance improvements of +42.3\% in missing value prediction, +79.1\% in dependency analysis, and +28.9\% in automated planning compared to baseline methods, showcasing its potential to revolutionize construction workflows and inspire domain-specific LLM advancements.

\end{abstract}

\begin{figure*}[ht]
    \centering
    \includegraphics[width=\linewidth]{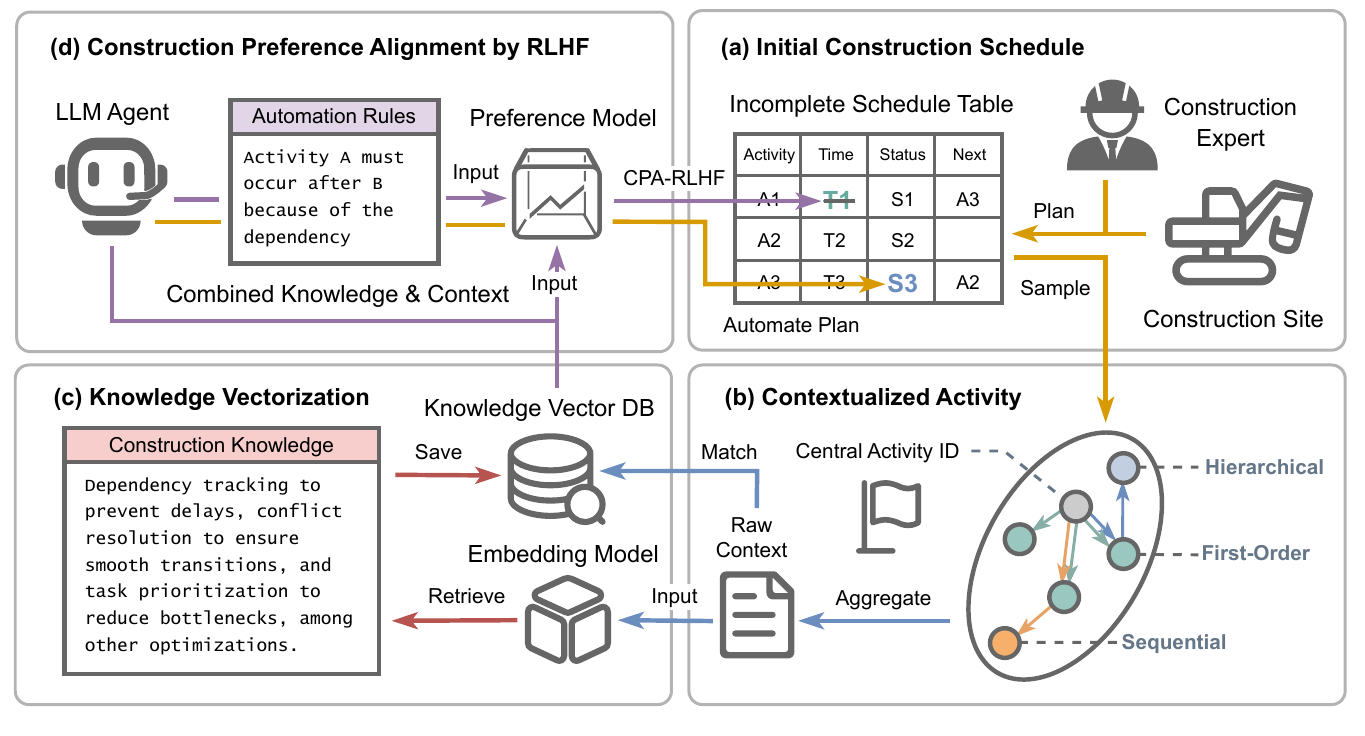}
    \caption{Overview of the \TheName{} system. (a) The initial construction schedule is created by experts and refined with contextual activity and site samples. (b) Contextualized activity aggregates hierarchical, first-order, and sequential relations. (c) Knowledge vectorization embeds and retrieves construction knowledge for optimization. (d) Construction preference alignment uses RLHF to align schedules with expert rules and preferences.}
    \label{fig:constructa_overview}
\end{figure*}

\section{Introduction}

Automating construction schedules in large-scale commercial projects, such as semiconductor fabrication, is an inherently complex task due to the dynamic nature of project contexts, intricate dependency structures, and the critical need for expert-driven decision-making~\cite{neelamkavil2009automation,azimi2011framework}. The difficulty lies in managing the vast number of interdependent activities, each with unique resource requirements and constraints, while simultaneously adapting to real-time changes and unforeseen disruptions~\cite{zavadskas2004multicriteria}. These factors necessitate seamless integration of domain knowledge and human expertise to ensure project feasibility and efficiency. Traditional methods, relying on rigid rules and static assumptions, often fail to adapt to the variability and uncertainty inherent in large-scale construction projects, leaving a critical need for more flexible and context-aware approaches~\cite{alegre2016engineering,al2020uncertainty}.


Despite recent advancements in machine learning, the potential of large language models (LLMs) for construction scheduling remains underexplored due to several limitations. LLMs, pretrained on broad datasets, lack the domain-specific knowledge needed for intricate project dependencies and constraints~\cite{xu2024llm,banerjee2024cps}. Moreover, the size and complexity of construction plans make it impractical to load entire projects into LLMs for automation~\cite{gidado1996project}. Instead, construction scheduling demands dynamic handling of real-time updates and evolving conditions. LLMs face three key challenges: (1) capturingthe intricate dependencies between construction activities, (2) adapting to context-sensitive changes in task priorities or resource availability, and (3) aligning outputs with expert-driven preferences. These challenges highlight the need for tailored frameworks to bridge the gap between LLM capabilities and the demands of large-scale construction projects.


To address these limitations, we present \TheName{}\footnote{\TheName{} and Construction RLHF are used interchangeably in this paper.}, a novel framework designed to optimize construction schedules dynamically by leveraging LLMs with three key components: (1) Static Retrieval-Augmented Generation (SRAG or Static RAG), which introduces domain-specific construction knowledge, enabling LLMs to understand definitions, rules, and constraints critical to commercial construction; (2) Contextualized Knowledge RAG~(Knowledge RAG or KRAG), which incorporates the expertise of architects by dynamically sampling context-sensitive information, ensuring the relevance of inputs to evolving project conditions; and (3) Construction RLHF, which aligns the outputs of LLMs with expert feedback to enhance their in-depth understanding and produce human-aligned scheduling decisions.

We evaluate \TheName{} on a proprietary dataset comprising 4,340 semiconductor fabrication activities characterized by intricate dependencies and constraints. \TheName{} delivers substantial performance improvements, including a 42.3\% boost in missing value prediction, 79.1\% in dependency analysis, and 28.9\% in automated planning compared to baseline methods. Further analysis across levels and areas shows adaptability, while Construction RLHF distills raw data into actionable insights, demonstrating scalability and robustness for complex construction tasks.

\begin{figure*}[htbp]
    \centering
    \includegraphics[width=\textwidth]{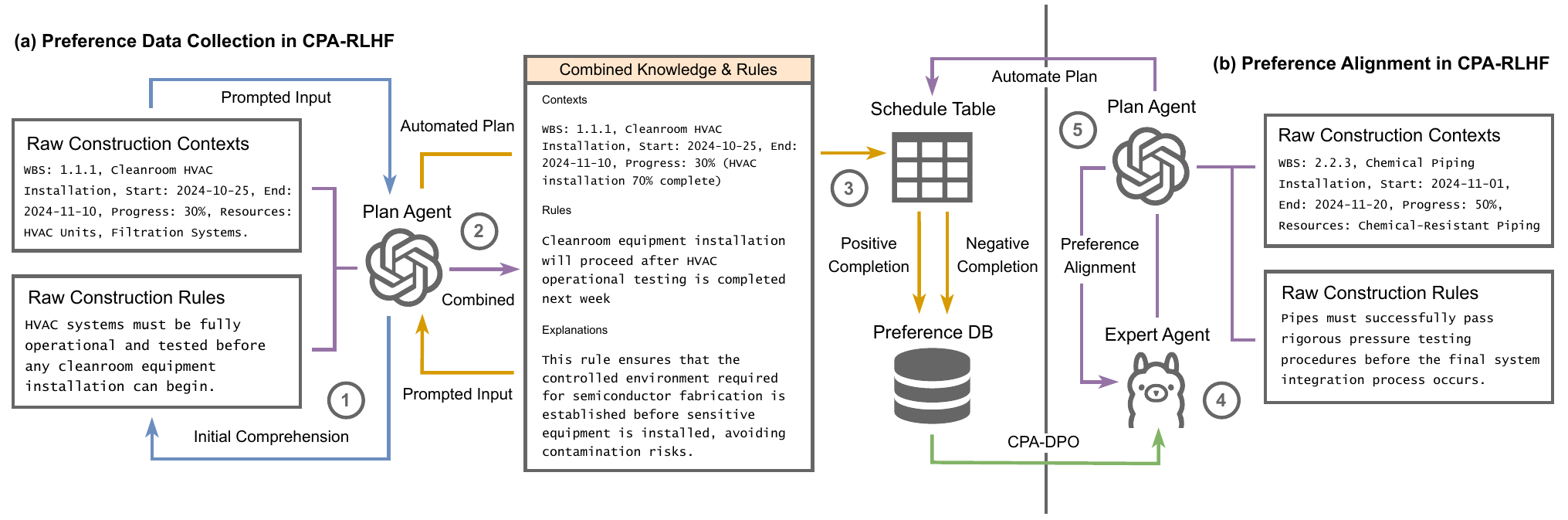}
    \caption{Illustration of the CPA-RLHF process. \textcircled{1} Raw contexts and rules are input for comprehension. \textcircled{2} The Plan Agent refines these into filtered contexts and rules. \textcircled{3} Completions are evaluated and stored in the Preference Database. \textcircled{4} The Expert Agent aligns outputs with project preferences. Part (a) collects data for preference model training, and part (b) aligns preferences for accurate planning.}
    \label{fig:constructa_cpa}
\end{figure*}

\section{Methodology}

Our methodology starts with an expert-provided schedule (Figure~\ref{fig:constructa_overview}, part (a)) and refines it using Static RAG for retrieval, Knowledge RAG for dependencies, and Construction RLHF for rule alignment (parts (c), (b), and (d)). The outputs, including retrieved knowledge and preference-aligned task relationships, are integrated into prompts for dynamic, context-aware scheduling.



\subsection{Static Retrieval-Augmented Generation}

Static RAG equips LLMs with construction-specific knowledge, as shown in part (c) of Figure~\ref{fig:constructa_overview}. It bridges the gap between general-purpose models and scheduling needs by generating embeddings for retrieval, with Local Static RAG providing precise definitions and Global Static RAG offering broader domain knowledge.

\textbf{Local Static RAG} provides precise definitions for construction-specific terms like Work Breakdown Structure (WBS) using curated online resources. For each term \( t \) in the terminology set \( \mathcal{T} \), its definition \( d_t \) is retrieved and embedded as \( e_t = f_\text{embed}(d_t) \) using an embedding model \( f_\text{embed} \). These embeddings are stored for contextualizing activities in schedule optimization.

\textbf{Global Static RAG} retrieves domain-specific knowledge from resources like textbooks or manuals. Raw text \( \mathcal{D} \) is cleaned and segmented into chunks \( \mathcal{C} = \{c_1, c_2, \dots, c_n\} \), each embedded as \( e_{c_i} = f_\text{embed}(c_i) \) and stored in a database. For a query \( q \), the system retrieves the most relevant chunk \( c^* \) by maximizing similarity \( \text{sim}(e_q, e_{c_i}) \), where \( e_q = f_\text{embed}(q) \) and \( c^* = \arg \max_{c_i \in \mathcal{C}} \text{sim}(e_q, e_{c_i}) \). Combining Local and Global Static RAG ensures precise definitions and broad domain knowledge for construction scheduling.

\subsection{Contextual Knowledge RAG}

Contextual Knowledge RAG samples task-specific contexts from a dependency graph \( G = (V, E) \), where \( V \) represents activities and \( E \) their dependencies. As shown in part (b) of Figure~\ref{fig:constructa_overview}, it aggregates hierarchical, first-order, and sequential relationships, using the combined context to retrieve relevant embeddings from the knowledge database for construction scheduling.

\textbf{Sequential Context} captures predecessor and successor activities up to three hops by traversing the graph in both directions. Random paths are sampled to reflect relevant sequential relationships while avoiding revisits and cycles, ensuring the selection of meaningful task flows.

\textbf{Hierarchical Context} retrieves nodes within the same Work Breakdown Structure (WBS) up to two levels. Tasks sharing WBS attributes are identified, and bidirectional traversal ensures that hierarchically consistent nodes are included in the context.

\textbf{First-Order Context} includes direct predecessors and successors of the target node, focusing on immediate task dependencies critical for accurate schedule representation.

Each task \( i \) is assigned a combined context \( C_i = \{\text{FirstOrder}(i), \text{Hierarchical}(i), \text{Sequential}(i)\} \), reflecting one-hop, two-hop, and three-hop constraints. Using the same embedding model as Static RAG, embeddings for \( C_i \) retrieve local knowledge and the top three global knowledge chunks from books and references, balancing dependencies to optimize rule generation and scheduling.

\subsection{Construction RLHF}

The Construction RLHF pipeline (Figure~\ref{fig:constructa_cpa}) refines schedules by integrating expert feedback and dynamic adjustments. Starting with raw contexts and rules (\textcircled{1}), the Plan Agent combines task-specific details with context retrieved from SRAG and KRAG (\textcircled{2}). Refined outputs, evaluated as positive or negative completions, are stored in the Preference Database (\textcircled{3}). The smaller Expert Agent\footnote{The dual-agent structure enables the smaller LLM to memorize preferences while the larger LLM automates schedules.}, compared to the Plan Agent, utilizes this feedback and memorized domain knowledge to ensure schedules align with dynamic project requirements (\textcircled{4}), supporting robust and adaptive scheduling.


\textbf{CPA-RLHF} acts as the overarching framework, transforming the initial construction schedule into a dynamic environment for offline reinforcement learning. This is achieved by masking certain ground-truth values to simulate real-world uncertainties, effectively leveraging the expertise of architects in providing feedback on schedule optimization. The masked environment serves as a feedback loop where evaluated completions inform the refinement of the preference model. This process enables CPA-RLHF to address complex scheduling requirements by integrating domain knowledge, contextual adjustments, and expert preferences.

Within this framework, \textbf{CPA-DPO} refines the preference alignment process through supervised fine-tuning (SFT) and direct preference optimization. SFT establishes an initial alignment by minimizing the cross-entropy loss \( L_{\text{SFT}} = -\frac{1}{N} \sum_{i=1}^{N} y_i \log p_i \), grounding the model in expert-labeled schedules to produce coherent and contextually relevant outputs. Building on this, the preference alignment phase optimizes the total loss \( L_{\text{total}} = L_{\text{SFT}} + \alpha L_{\text{CR}} + \beta L_{\text{PA}} \), where \( \alpha \) and \( \beta \) balance contributions from Context-Rule Interaction Loss (\( L_{\text{CR}} \)) and Preference Alignment Loss (\( L_{\text{PA}} \)). The latter, defined as \( L_{\text{PA}} = -\frac{1}{N} \sum_{i=1}^{N} \left( y_i \log(p_i) + (1 - y_i) \log(1 - p_i) \right) \), ensures model outputs align with expert-defined preferences while respecting project constraints. This integrated approach enables the model to dynamically adapt to construction complexities, improving task prioritization and resource allocation.

\section{Experimental Design}

This section outlines the experimental configurations for Static RAG, Knowledge RAG and Construction RLHF, emphasizing embedding methods, model configurations, and optimization strategies.

\textbf{Static and Knowledge RAG} The SRAG setup used 500-token chunks for efficient processing, with embeddings generated via \texttt{all-MiniLM-L6-v2}\footnote{\url{https://huggingface.co/sentence-transformers/all-MiniLM-L6-v2}}. Static RAG focused on terminologies and definitions, while Knowledge RAG retrieved context from manuals and domain-specific references.

\textbf{Construction RLHF} The Plan Agent used GPT-4o~\cite{islam2024gpt}, and the Expert Agent employed Llama3.2-3B model~\cite{touvron2023llama} for expert preference alignment. Training involved 10 epochs of SFT for initialization, followed by 10 epochs of CPA-DPO for preference refinement. The trained Expert Agent supported contextual refinements. 

\textbf{LLM Training Configuration} Efficient training was achieved using 4-bit quantization, gradient checkpointing, mixed precision training, and the AdamW optimizer~\cite{zhuang2022understanding}. Data collection employed a random seed of 42, while inference utilized a seed of 12345, ensuring the generation of diverse datasets to enhance generalizability.

\textbf{Prompt Design} Comprehensive prompt categories tailored for each task are provided in the appendix to address construction-specific challenges effectively. Each result reflects the top-2 predictions (\( k = 2 \)) for enhanced accuracy, with Construction RLHF ensembled with KRAG to combine expert alignment and domain-specific knowledge retrieval.

\section{Result and Analysis}

We evaluate \TheName{} across key scheduling tasks, highlighting its ability to address complex dependencies, handle missing data, and align schedules with expert-defined constraints.

\subsection{Evaluation Metrics}

\TheName{} is evaluated using three key metrics to assess its ability to predict missing elements in construction schedules while ensuring logical consistency and expert alignment.

\textbf{Missing Value Prediction (MVP)} measures the model’s ability to reconstruct values from three randomly removed columns. This tests its capability to handle incomplete data while preserving schedule coherence and minimizing disruptions caused by missing information.

\textbf{Dependency Analysis (DA)} evaluates prediction accuracy for relational columns, including Activity Status, Level, Area, and Discipline. Since these dependencies define task sequencing and workflow constraints, this metric ensures that predicted schedules maintain logical task relationships and prevent inconsistencies.

\textbf{Automated Planning (AP)} assesses the model’s ability to predict Current Start and Current Finish dates while considering real-world constraints. It measures how well the generated schedules align with expert workflows, resource availability, and project feasibility to ensure practical execution.

\begin{table}[t]
\centering
\resizebox{\columnwidth}{!}{
\begin{tabular}{l|ccc|c}
\toprule
Model Config            & MVP (\%)         & DA (\%)         & AP (\%)         & Avg (\%) \\ \midrule
GPT-4o~(Basic Context)  & 14.6             & 3.1             & 8.4             & 8.7      \\ \midrule
+ Static RAG            & 11.6             & 1.6             & 12.5            & 8.6      \\
+ Knowledge RAG         & 51.4             & 77.9            & 25.9            & 51.7     \\
+ Construction RLHF     & 56.9             & 82.2            & 37.3            & 58.8     \\ \midrule
Gain (\TheName{} vs. BC) & \textbf{+42.3}  & \textbf{+79.1}  & \textbf{+28.9}  & \textbf{+50.1}  \\ 
\bottomrule
\end{tabular}
}
\caption{Performance comparison of pretraining configurations for construction schedule optimization. 
\textbf{Basic Context (BC)} refers to GPT-4o without retrieval augmentation or RLHF, relying only on general pretraining knowledge by sampling random rows as context.}
\label{tab:ablation_study_construction}
\end{table}

\begin{table*}[h]
\centering
\resizebox{\textwidth}{!}{
\begin{tabular}{l|l|cccc|cccc|cccc}
\toprule
Group & Discipline & \multicolumn{4}{c|}{MVP (\%)} & \multicolumn{4}{c|}{DA (\%)} & \multicolumn{4}{c}{AP (\%)} \\ 
 &  & BC & SRAG & KRAG & RLHF & BC & SRAG & KRAG & RLHF & BC & SRAG & KRAG & RLHF \\ \midrule
CSA & CSA.Arch.Arch-D & 5.6 & 4.4 & 23.3 & 25.6 & 1.7 & 2.5 & 39.2 & 43.3 & 0.8 & 5.0 & 15.0 & 19.2 \\
 & CSA.Arch.CRCs-D & 6.7 & 6.7 & 40.0 & 40.0 & 0.0 & 0.0 & 65.0 & 65.0 & 0.0 & 0.0 & 17.5 & 20.0 \\
 & CSA.Arch.Metal & 6.5 & 4.5 & 32.7 & 37.8 & 2.1 & 0.7 & 61.0 & 64.2 & 3.1 & 7.5 & 10.8 & 16.8 \\
 & CSA.Arch.RF & 9.5 & 6.3 & 38.1 & 42.9 & 0.9 & 0.6 & 49.7 & 53.3 & 2.1 & 6.0 & 8.0 & 17.3 \\ 
 & CSA.Arch.WPRF & 0.0 & 0.0 & 33.3 & 33.3 & 0.0 & 0.0 & 25.0 & 25.0 & 0.0 & 25.0 & 0.0 & 0.0 \\ 
 & CSA.Civil.Earthwork & 11.3 & 6.9 & 32.8 & 38.2 & 1.6 & 0.4 & 47.2 & 52.0 & 3.6 & 6.9 & 10.1 & 16.5 \\ 
 & CSA.Struc.Concrete & 8.8 & 7.4 & 29.2 & 33.0 & 1.1 & 1.3 & 35.5 & 39.1 & 4.6 & 6.7 & 12.9 & 20.0 \\ 
 & CSA.Struc.Modules & 6.2 & 5.6 & 37.1 & 41.7 & 2.7 & 0.5 & 61.8 & 66.5 & 3.8 & 6.9 & 11.6 & 19.7 \\
 & CSA.Struc.Piers & 7.5 & 6.7 & 30.0 & 36.7 & 0.0 & 0.0 & 45.0 & 47.5 & 7.5 & 5.0 & 22.5 & 32.5 \\
 & CSA.Struc.Steel & 8.0 & 7.8 & 30.8 & 34.0 & 1.7 & 1.0 & 50.3 & 53.6 & 3.5 & 7.4 & 15.4 & 21.6 \\ 
 & CSA.Struc.Strut & 9.0 & 5.6 & 33.5 & 37.9 & 1.8 & 0.8 & 60.1 & 64.9 & 6.5 & 5.8 & 18.2 & 27.1 \\
 \midrule
MEP & MEP.Mech.Dry & 4.6 & 6.4 & 37.6 & 38.8 & 2.5 & 0.4 & 65.7 & 68.0 & 4.6 & 6.4 & 18.2 & 25.4 \\
 & MEP.Mech.Wet & 0.0 & 0.0 & 66.7 & 66.7 & 0.0 & 0.0 & 75.0 & 75.0 & 0.0 & 0.0 & 50.0 & 50.0 \\ 
 & MEP.Proc.HP & 3.2 & 5.4 & 33.3 & 40.8 & 1.4 & 0.2 & 61.5 & 66.5 & 3.2 & 5.4 & 14.0 & 23.7 \\ 
 & MEP.Proc.LP & 4.3 & 4.8 & 35.2 & 39.3 & 1.6 & 0.5 & 61.1 & 68.2 & 4.3 & 6.9 & 14.6 & 22.9 \\ 
 & MEP.Proc.Vac & 7.5 & 5.2 & 32.4 & 36.7 & 1.1 & 0.7 & 57.5 & 63.9 & 7.5 & 6.8 & 19.6 & 31.1 \\ 
 & MEP.Proc.Waste & 7.9 & 6.4 & 33.1 & 38.8 & 1.4 & 0.4 & 63.0 & 67.9 & 5.2 & 6.8 & 18.2 & 25.2 \\ 
 & MEP.Proc.Water & 5.7 & 6.4 & 37.6 & 38.8 & 3.0 & 0.5 & 65.7 & 70.5 & 3.8 & 7.6 & 14.8 & 23.2 \\
 \midrule
 Avg & & 6.2 & 5.2 & 36.4 & 40.2 & 1.5 & 0.6 & 55.8 & 59.2 & 3.7 & 6.6 & 17.4 & 24.3 \\
\bottomrule
\end{tabular}
}
\caption{Grouped performance comparison across construction schedule optimization tasks. 
SRAG retrieves domain-specific definitions, KRAG structures context using activity relationships, and RLHF aligns predictions with expert feedback. Results show notable gains in MVP, DA, and AP, especially in CSA and MEP disciplines.}
\label{tab:cross_column_ablation_study_with_rlhf_full}
\end{table*}



\subsection{Overall Performance Gains}

Table~\ref{tab:ablation_study_construction} demonstrates the overall performance improvements of \TheName{} across MVP, DA, and AP tasks. Static RAG shows limited impact, with marginal or decreased performance, as it provides domain knowledge without contextual adaptation. Knowledge RAG boosts MVP and DA by incorporating task-specific dependencies, improving inference of missing values and logical sequencing. Construction RLHF achieves the highest gains, improving MVP by +42.3\%, DA by +79.1\%, and AP by +28.9\%. These results highlight the effectiveness of \TheName{} in addressing complex construction scheduling tasks.


\begin{figure*}[h]
    \centering
    \includegraphics[width=\textwidth]{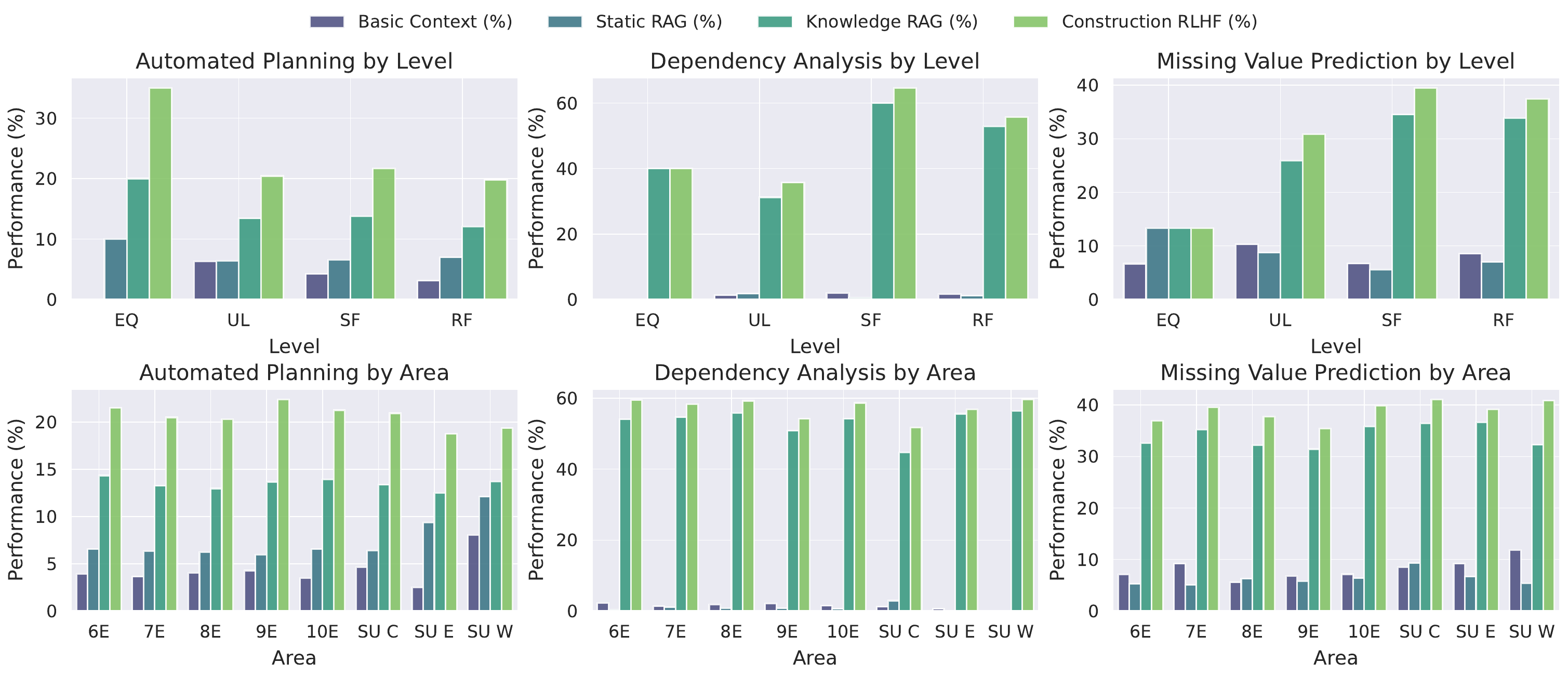}
    \caption{Performance Comparison Across Levels and Areas. This plot shows the performance of various metrics, including Basic Context, Static RAG, Knowledge RAG, and Construction RLHF, for three tasks (Automated Planning, Dependency Analysis, and Missing Value Prediction) across different levels and areas.}
    \label{fig:performance_comparison}
\end{figure*}

\begin{figure*}[h]
    \centering
    \includegraphics[width=\textwidth]{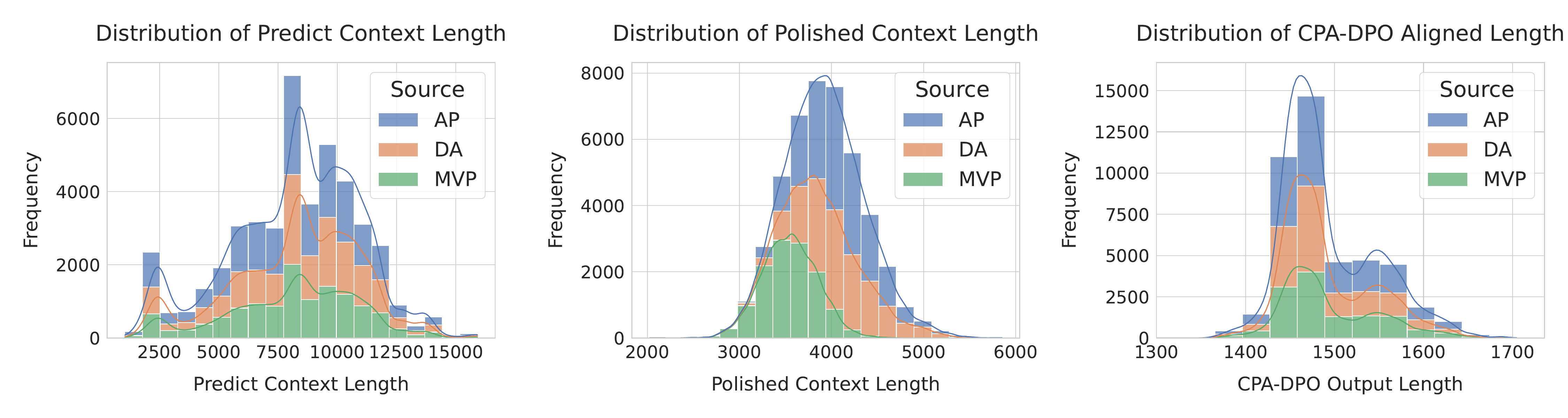}
    \caption{Context length distributions for AP, DA, and MVP sources, highlighting reductions achieved through CPA-DPO. The shorter contexts effectively maintain performance while improving efficiency in schedule optimization.}
    \label{fig:distribution_by_source}
\end{figure*}

\subsection{Construction Disciplines, Levels, and Areas in Evaluation}

Effective construction scheduling depends on disciplines, structural levels, and spatial areas, each with unique dependencies. We evaluate \TheName{} across these dimensions to ensure adaptability to real-world constraints.

\textbf{Disciplines} Construction projects encompass Civil, Structural, and Architectural (CSA) and Mechanical, Electrical, and Plumbing (MEP) disciplines. CSA tasks, such as structural assemblies and load-bearing elements, require precise sequencing for stability. MEP tasks, including waste processing and high-pressure systems, demand coordinated integration for efficient infrastructure.  

\textbf{Levels} Evaluation covers Equipment (EQ), Utility Level (UL), Standard Floor (SF), and Roof Floor (RF). SF and RF are the most complex, with RF requiring detailed sequencing for reinforcements and installations.  

\textbf{Areas} Performance is analyzed in construction zones such as 6E, 9E, and SU. High-complexity areas like SU E and 10E have dense interdependencies, making effective scheduling essential for coordination and resource optimization.  





\subsection{Performance by Discipline}

The grouped results in Table~\ref{tab:cross_column_ablation_study_with_rlhf_full} provide insights into \TheName{}’s performance across construction disciplines. For CSA disciplines, including CSA.Struc.Modules and CSA.Struc.Piers, \TheName{} excels in accurately modeling dependencies and generating optimized schedules, effectively addressing challenges such as sequencing structural assemblies, ensuring load-bearing integrity, and maintaining alignment with construction constraints. 

Similarly, for MEP disciplines, including MEP.Proc.Waste and MEP.Proc.HP, significant improvements are observed in DA and AP, demonstrating \TheName{}’s ability to capture intricate interdependencies between mechanical, electrical, and plumbing systems. This highlights the model’s robustness in specialized workflows where precise coordination of installations and operational constraints is critical to overall project efficiency.



\subsection{Performance by Level and Area}

Figure~\ref{fig:performance_comparison} compares performance across construction levels (EQ, UL, SF, RF) and areas (6E, 9E, SU). \TheName{} consistently outperforms other methods across all categories, demonstrating its ability to adapt to varying spatial and structural complexities. 

For levels, the largest improvements are observed in SF and RF, highlighting the model's capability to handle complex roof-level dependencies, structural reinforcements, and standard floor operations with greater accuracy. The gains in RF indicate that \TheName{} effectively accounts for elevated sequencing constraints and installation workflows that are more intricate at higher levels.

For areas, \TheName{} achieves the highest gains in zones with high complexity, such as SU E and 10E, where interdependencies between tasks are more intricate. This suggests that \TheName{} effectively learns and adapts to localized construction constraints, optimizing sequencing and resource allocation in highly constrained or densely coordinated zones.



\subsection{Knowledge Distillation and Observations}

Figure~\ref{fig:distribution_by_source} shows the reduced context length after CPA-DPO alignment, demonstrating effective knowledge distillation from the Plan Agent to the Expert Agent. By filtering out redundant details and retaining only essential scheduling constraints, \TheName{} enhances efficiency while preserving decision-making accuracy. By prioritizing critical dependencies, it enables more precise scheduling adjustments and minimizes the risk of misaligned task sequencing.


Another key observation is that \TheName{} refines scheduling inputs by reducing context length while preserving essential constraints. CPA-DPO alignment streamlines DA and MVP, filtering excess details that obscure dependencies. This distillation enhances adaptability by emphasizing key relational structures, improving interpretability and alignment with industry requirements.


\section{Future Applications and Industry Adoption}

\TheName{} presents strong potential for LLM adoption in construction scheduling, improving automation, adaptability, and decision support. Traditional methods struggle with real-time changes, while \TheName{} continuously refines schedules based on evolving constraints~\cite{pan2021automated,neelamkavil2009automation}. By learning from historical schedules and domain-specific constraints, it optimizes resource allocation, mitigates conflicts, and enhances project execution.

For broader adoption, \TheName{} can integrate with existing construction management software as an intelligent planning tool. Its ability to handle dynamic scheduling and dependency modeling makes it valuable for large-scale projects. Future work will address deployment challenges, including computational efficiency, latency, and seamless integration with industry platforms~\cite{zhang2023rule,amer2023construction}, ensuring scalability for commercial applications such as semiconductor fabrication.

\section{Related Works}

Research on LLM-powered construction scheduling is limited, with prior work focusing on deterministic methods and RL in other domains~\cite{srivastava2022imperative,dashti2021integrated,bademosi2021factors,pan2021automated,li2021optimal}. This work pioneers construction automation using RAG and RLHF.



\subsection{Construction Automation}

Traditional construction automation has predominantly utilized deterministic scheduling algorithms~\cite{peiris2023production,khodabakhshian2023deterministic,peiris2023production} and rule-based systems~\cite{zhang2023rule,amer2023construction,augar2024rule}. While these methods are effective in static environments, they often fail to adapt to the dynamic and complex nature of real-world construction projects, which involve evolving dependencies and resource constraints~\cite{xie2023case,al2024generation,parekh2024automating,he2024real,huang2024cross}. Our approach addresses these limitations by integrating domain-specific knowledge and context, enabling more flexible and responsive scheduling.

\subsection{Retrieval-Augmented Generation~(RAG)}

Retrieval-Augmented Generation (RAG) techniques enhance language models by incorporating external knowledge sources, improving their ability to generate contextually relevant information~\cite{gao2023retrieval,chen2024benchmarking,jiang2024longrag,li2024malmixer,acharya2025optimizing}. However, existing RAG methods may not effectively retrieve and integrate the highly specialized and structured information required for construction scheduling~\cite{zhao2024retrieval,fan2024survey,barnett2024seven}. Our method overcomes this challenge by employing a static RAG framework tailored to the construction domain, ensuring the retrieval of precise and pertinent information that informs scheduling decisions.



\subsection{Reinforcement Learning from Human Feedback~(RLHF)}

Reinforcement Learning from Human Feedback (RLHF), including Direct Preference Optimization (DPO), aligns model outputs with human preferences through comparative feedback~\cite{wang2023rlhf,yang2024multi,dong2024rlhf,xu2024dpo,saeidi2024insights}. In software engineering, RLHF has been used to enhance model alignment with human reasoning, leveraging human attention and feedback to improve code summarization, model focus, and explainability~\cite{bansal2023modeling,karas2024tale,li2024machines,zhang2024eyetrans}. Additionally, studies show that LLMs can learn structured decision patterns from human-provided code comments and summarization patterns~\cite{zhang2022pre,zhang2022leveraging}, demonstrating RLHF’s potential for domains requiring contextual understanding, such as construction.


However, applying RLHF in traditional industries like construction remains challenging due to the need for domain-specific knowledge, complex dependencies, and expert-driven priorities~\cite{wang2024comprehensive,xiao2024comprehensive,feng2024towards}. While RLHF has been applied in various domains, its use in construction scheduling remains underexplored. Our approach extends DPO by incorporating construction-specific knowledge and structured context, resulting in schedules that better reflect expert preferences and project-specific requirements.

\section{Conclusion}

In conclusion, we presented \TheName{}, an approach for automating construction schedules by integrating LLMs, contextualized knowledge RAG, and RLHF to optimize workflows with expert input. This framework advances traditional methods, offering flexibility, scalability, and adaptability for large-scale projects with complex dependencies. Future work includes implementing the Construction DPO model, incorporating multimodal inputs, and evolving \TheName{} into a dynamic recommender system for continuous project adaptation.

\section*{Acknowledgment}

This work was supported by Intel Corporation\footnote{\url{https://www.intel.com/content/www/us/en/homepage.html}}, specifically through the Incubation and Disruptive Innovation group. We also appreciate the collaboration and insights from Intel Foundry\footnote{\url{https://www.intel.com/content/www/us/en/foundry/overview.html}} employees, whose expertise in semiconductor fabrication has guided our exploration of leveraging LLMs to automate construction processes in chip manufacturing.

\bibliography{anthology}

\appendix
\section*{Appendix: Additional Details}
\renewcommand{\thesubsection}{A.\arabic{subsection}} 

In this appendix, we provide comprehensive details on the experiments conducted, including sensitivity analysis on context embedding models, variations of preference alignment strategies, the complexity analysis of the construction dependency graph, and the detailed design of context sampling methods, prompt categories, and task-specific prompts.

\subsection{Complexity of the Construction Dependency Graph}

\begin{figure}[t]
    \centering
    \includegraphics[width=\columnwidth]{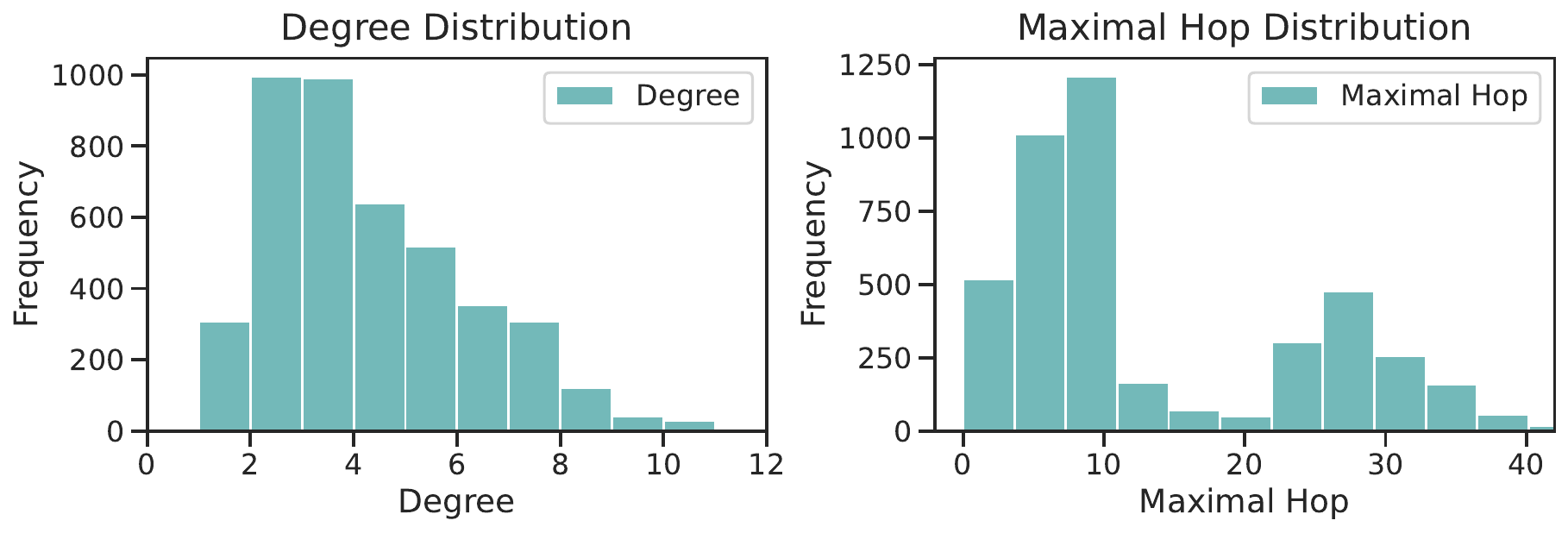}
    \caption{Distribution of degree and maximal hop for dependency graph nodes. The left plot shows the degree distribution, reflecting task interconnectivity, while the right plot presents the maximal hop distribution, highlighting long-range task dependencies.}
    \label{fig:degree_hop_distribution}
\end{figure}

Understanding the structural complexity of the dependency graph is critical for automating construction schedules effectively. We analyzed two key metrics to highlight the challenges posed by real-world construction scenarios (Figure~\ref{fig:degree_hop_distribution}):

\begin{itemize}
    \item \textbf{Degree Distribution}: This metric captures the number of connections each activity node has within the dependency graph. As shown in Figure~\ref{fig:degree_hop_distribution}, the degree distribution exhibits a mean value of 3.86, with some nodes having as many as 20 connections. These values indicate the extensive interdependencies among activities, which require careful management to maintain project feasibility and avoid resource bottlenecks.
    
    \item \textbf{Maximal Hop Distribution}: This measures the farthest distance, in terms of hops, to dependent nodes. The average maximal hop distance is 13.93, with the highest value reaching 73. These long-range dependencies demonstrate the need for multi-level propagation strategies to capture hierarchical and sequential task relationships effectively.
\end{itemize}

\noindent These metrics emphasize the intricate nature of construction scheduling, with both high interconnectivity and significant multi-level dependencies. The insights derived from these analyses underline the importance of advanced frameworks like \TheName{} to manage such complexity in commercial construction projects.

\subsection{Correlation and Similarity Analysis of Project Attributes}

Understanding relationships among project attributes is essential for optimizing construction scheduling and dependency management. We conducted two types of analyses to capture both linear correlations and deeper semantic relationships:

\begin{itemize}
    \item \textbf{Correlation Analysis (Encoded Data)}: We examined linear dependencies between attributes by encoding categorical data as numeric codes and calculating Pearson correlation coefficients across project attributes. This method identifies direct dependencies that impact the project timeline and resource allocation, revealing structural insights into task sequences.
    
    \item \textbf{Cosine Similarity Analysis (Embeddings)}: Using embeddings generated from the \texttt{distilbert-base-uncased}\footnote{\url{https://huggingface.co/distilbert-base-uncased}}
 pre-trained language model, we captured semantic relationships among attributes that linear correlations might miss. This analysis highlights implicit, context-driven dependencies such as role interactions and spatial relationships, providing a nuanced view of project structure.
\end{itemize}

\begin{figure*}[h]
    \centering
    \includegraphics[width=\textwidth]{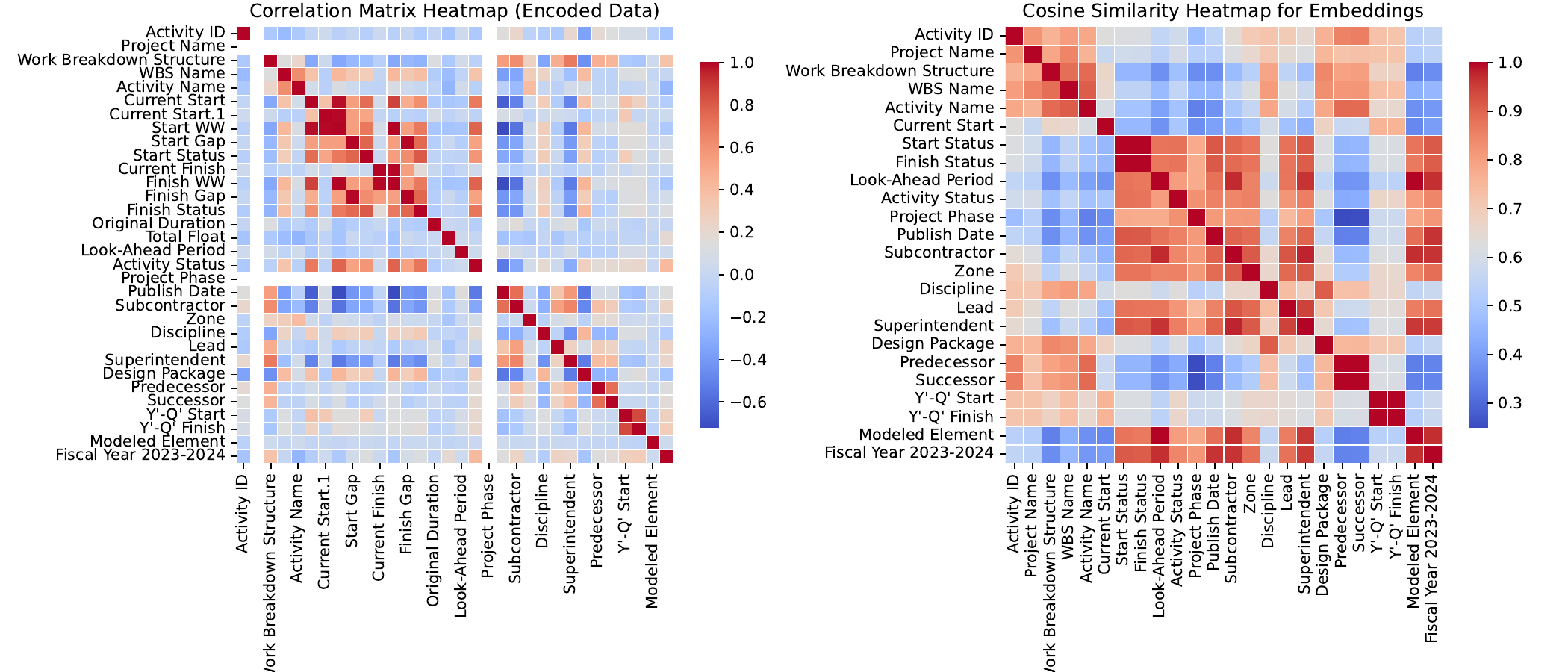}
    \caption{Combined Correlation and Cosine Similarity Heatmaps for Project Attributes. The left plot illustrates the correlation matrix based on encoded project data, highlighting linear relationships among attributes. The right plot presents the cosine similarity matrix based on embeddings, revealing deeper semantic associations among attributes.}
    \label{fig:combined_similarity_matrices}
\end{figure*}

\noindent Figure \ref{fig:combined_similarity_matrices} displays the results from both analyses, each providing unique insights:

\textbf{Correlation Matrix (Encoded Data)}: The left heatmap highlights linear relationships among attributes, with several notable correlations:

\begin{itemize}
    \item \textbf{Current Start and Current Finish}: The high correlation here reflects the dependency between start and finish dates, a foundational aspect of project scheduling.
    
    \item \textbf{Activity Status and Project Phase}: Correlations between activity status and project phase suggest that certain statuses align with specific phases, informing phase-based scheduling prompts.
    
    \item \textbf{Predecessor and Successor}: Strong correlation indicates that tasks have sequential dependencies, essential for creating an accurate task sequence.
\end{itemize}

In summary, these correlations reveal structural dependencies in project attributes, assisting in identifying key points in the scheduling and sequencing workflow. These insights enable more effective scheduling strategies by understanding which attributes inherently impact each other.

\textbf{Cosine Similarity Matrix (Embeddings)}: The right heatmap reveals semantic relationships between attributes, which help identify context-based dependencies:

\begin{itemize}
    \item \textbf{Subcontractor and Superintendent}: High similarity implies overlapping responsibilities between these roles, which can guide role-based dependencies in scheduling.
    
    \item \textbf{Discipline and Zone}: This similarity reflects the association between certain disciplines and zones, useful for location-based dependency prompts.
    
    \item \textbf{Project Phase and Activity Status}: Semantic alignment between phases and statuses provides a structured basis for task progression, useful for designing prompts that ensure coherent task sequences.
\end{itemize}

Overall, these embedding-based relationships uncover context-driven dependencies beyond simple correlations, offering a richer view of the project structure. Such insights are critical for tasks involving nuanced scheduling needs, as they reveal role interactions and locational dependencies that direct scheduling and resource assignment decisions.

\subsection{Unified Context Sampling Visualization}

\begin{figure*}[ht]
    \centering
    \includegraphics[width=\textwidth]{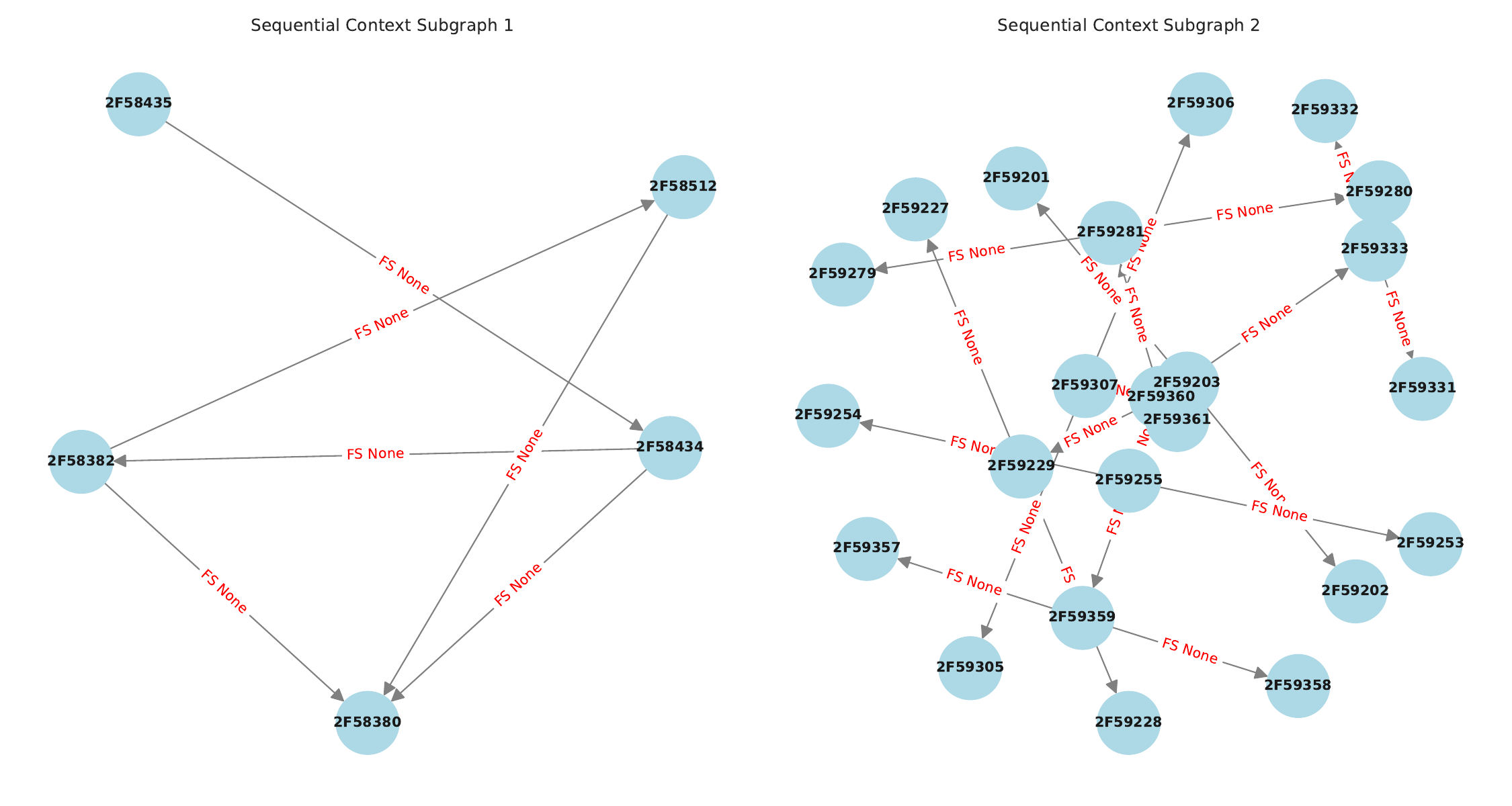}
    \caption{Sequential Context Sampling: This sampling method extracts nodes up to three hops away from each selected activity, representing predecessors and successors. Sequential sampling highlights dependencies that span across multiple stages in the construction workflow, enabling the model to understand task sequences and critical paths that influence the overall project schedule.}
    \label{fig:sequential_context}
\end{figure*}

\begin{figure*}[ht]
    \centering
    \includegraphics[width=\textwidth]{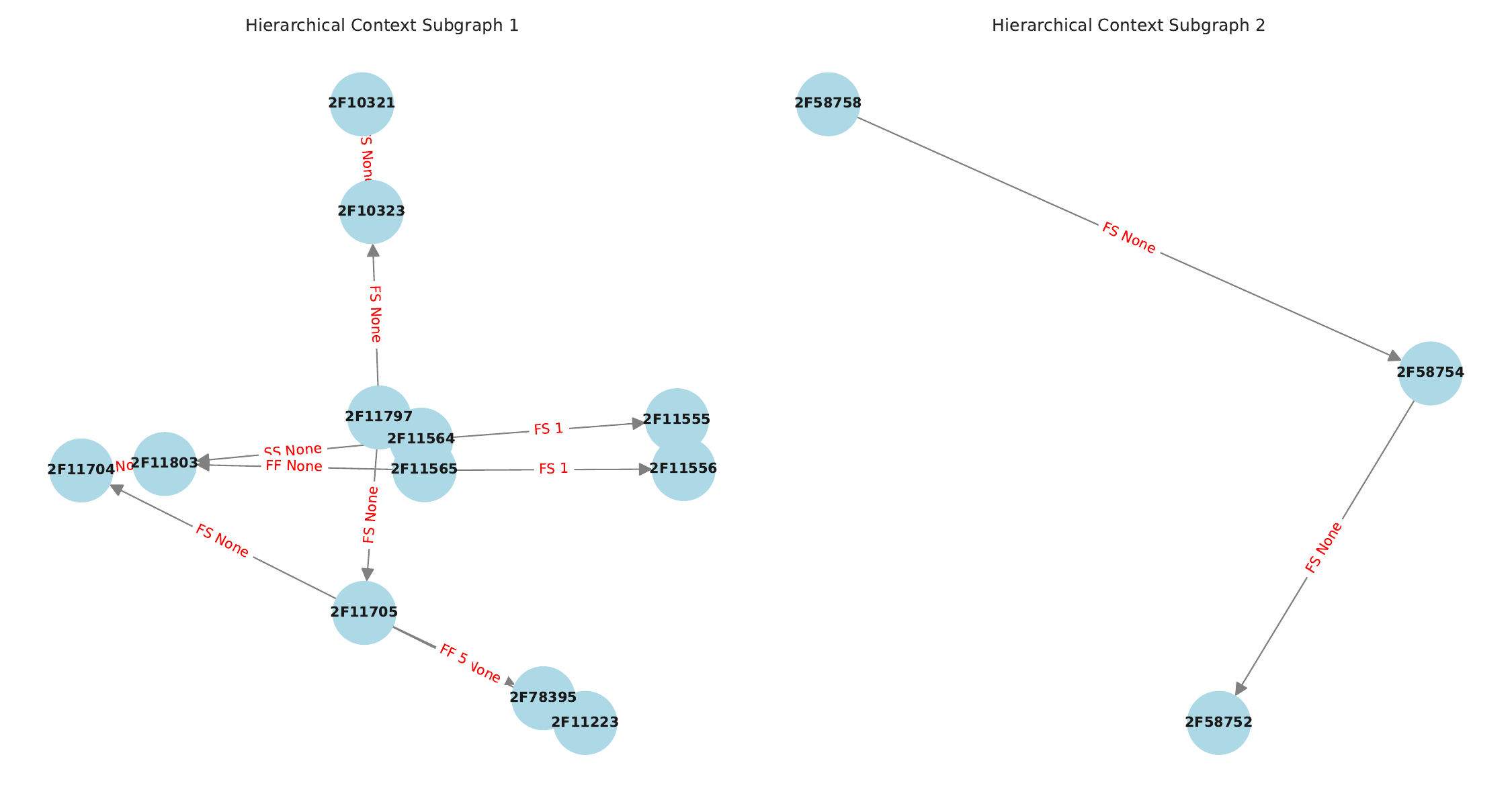}
    \caption{Hierarchical Context Sampling: This sampling focuses on capturing nodes within the same Work Breakdown Structure (WBS) up to two hops. Hierarchical context provides insights into tasks grouped by project phases, illustrating how dependencies within each WBS segment affect the schedule’s progression.}
    \label{fig:hierarchical_context}
\end{figure*}

\begin{figure*}[ht]
    \centering
    \includegraphics[width=\textwidth]{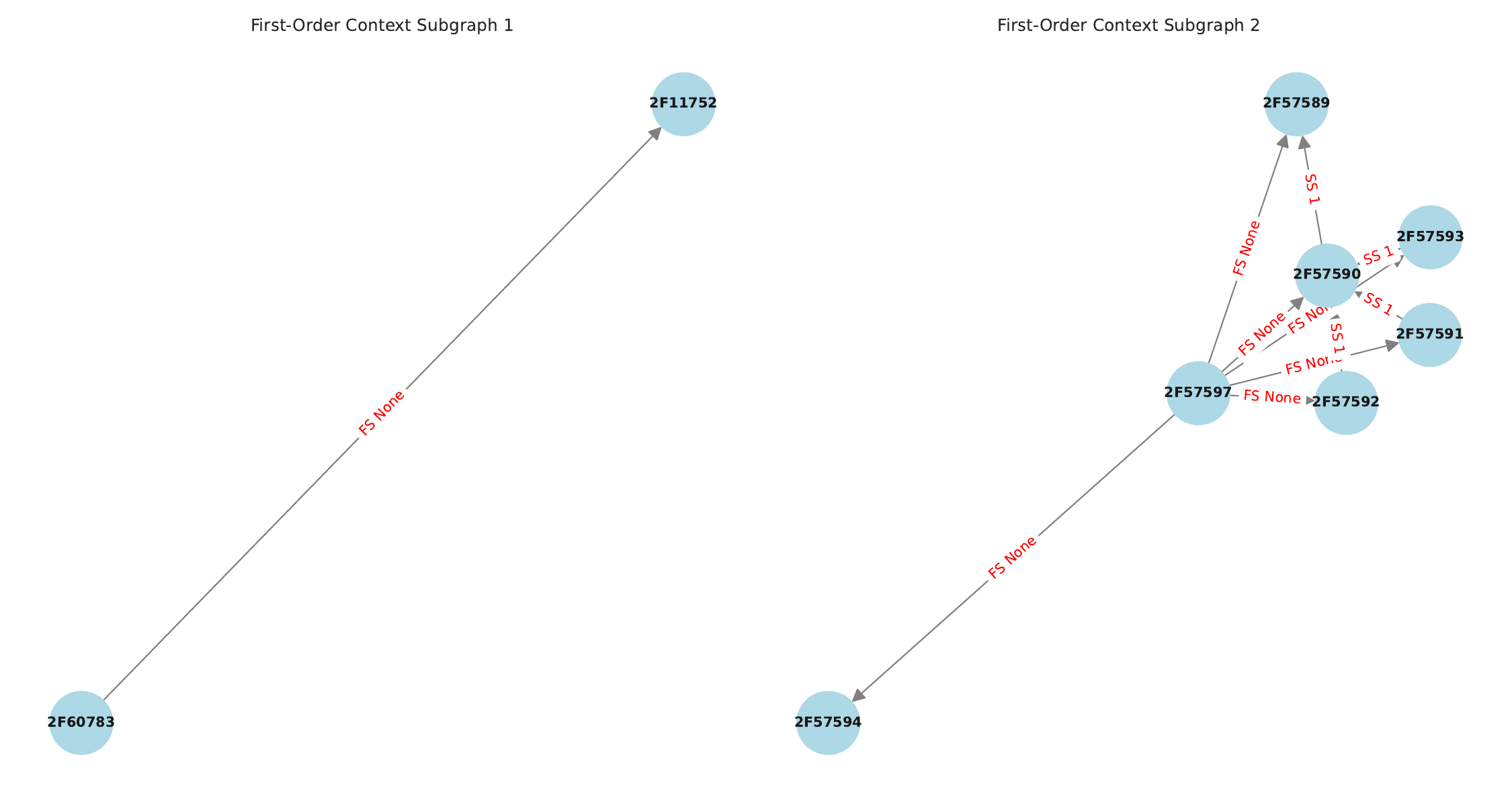}
    \caption{First-Order Context Sampling: This method captures only direct predecessors and successors for each selected activity. First-order context highlights immediate task dependencies, providing a concise view of direct task relationships essential for high-priority scheduling adjustments.}
    \label{fig:first_order_context}
\end{figure*}

To support effective construction scheduling, we employ a unified sampling method that extracts three distinct types of contextual information from project activities: Sequential Context, Hierarchical Context, and First-Order Context. Each method offers a unique approach to capturing dependencies and relationships among construction activities, facilitating comprehensive schedule optimization. Figures \ref{fig:sequential_context}, \ref{fig:hierarchical_context}, and \ref{fig:first_order_context} illustrate the structure and details of each context sampling method.

In Figure \ref{fig:sequential_context}, Sequential Context Subgraph 1 (left) shows a network of activities where nodes represent individual tasks required for project completion, connected by directed edges that denote task dependencies. Each node connects to predecessors and successors up to three hops away, capturing dependencies such as Finish-to-Start (FS), Finish-to-Finish (FF), Start-to-Start (SS), and, though less common, Start-to-Finish (SF) relationships. This structure is critical for visualizing the overall task flow, identifying critical paths, and highlighting potential bottlenecks that could delay project delivery. Sequential Context Subgraph 2 (right) extends this by including a larger set of interconnected nodes, where tasks are annotated with additional details such as task duration, resource requirements, and start or finish times. This dense layout offers a comprehensive view of task sequences, helping project managers forecast delays, pinpoint bottlenecks, and dynamically adjust schedules to accommodate unforeseen changes.

Figure \ref{fig:hierarchical_context} shows the Hierarchical Context Sampling. Hierarchical Context Subgraph 1 (left) presents nodes representing major project phases or milestones and their sub-tasks, organized within a structured hierarchy. Starting from a root node that signifies the overall project, dependencies cascade down through the graph, capturing relationships such as Start-to-Start and Finish-to-Start within a single WBS segment. This layout allows for visualizing dependencies specific to each phase, which is crucial for managing resources and time within discrete project stages. Hierarchical Context Subgraph 2 (right) shows a more streamlined arrangement, where tasks follow a linear progression, emphasizing phase-aligned scheduling adjustments. This structure helps project managers identify the critical path within each phase and adjust scheduling as needed to optimize workflow and resource allocation, while ensuring flexibility to adapt to phase-specific constraints and objectives.

Figure \ref{fig:first_order_context} illustrates the First-Order Context Sampling. First-Order Context Subgraph 1 (left) shows a minimal structure with only one dependency, representing a direct, Finish-to-Start relationship between two tasks. This sparse setup allows for focused adjustments on critical dependencies without the complexity of additional nodes, making it ideal for high-priority scheduling where immediate, direct task relationships are paramount. First-Order Context Subgraph 2 (right) presents a more intricate structure with multiple tasks directly connected to a central node. This setup captures immediate predecessors and successors, including Start-to-Start (SS) and Finish-to-Finish (FF) dependencies, providing a concise overview of key relationships around the central task. Such a layout enables project managers to address dependencies that directly impact the timing and prioritization of essential tasks, helping maintain schedule adherence while focusing on high-impact areas of the project.

Each sampling method uniquely extracts relevant information from the project table, allowing the model to adaptively balance broad, phase-level dependencies with immediate task relationships. This unified approach to context sampling is instrumental in generating a well-rounded understanding of the construction schedule, enabling dynamic and context-aware adjustments.

\subsection{General Predefined Prompt Categories and Context Mapping}

The prompt system utilizes predefined categories and context mappings to structure data collection for various tasks in construction scheduling. Each category aligns with specific aspects of project analysis, guiding the language model to interpret context effectively. This design ensures the capture of dependencies, durations, and resource-based relationships essential for scheduling. 

\begin{itemize}
    \item \textbf{Activity Sequence and Timing}: This prompt helps the model list construction activities based on 'Current Start' and 'Current Finish' dates, following dependencies defined by 'Predecessor Details' and 'Successor Details'. This captures the linear progression of tasks, aiding structured timeline generation.

    \item \textbf{Calculate Activity Duration}: Focusing on each activity's duration based on start and finish dates, this prompt aids in establishing a timeline for the project. The model uses these durations to enhance scheduling precision and identify critical periods in the workflow.

    \item \textbf{Hierarchical Tree Structure}: By organizing tasks according to the Work Breakdown Structure (WBS), this prompt helps arrange tasks hierarchically and identify sequential requirements, essential for maintaining the logical flow within each project phase.

    \item \textbf{Assess Sequence Reconstruction}: This prompt directs the model to assess if task sequences can be reconstructed from available data, highlighting missing elements. Such reconstruction ensures dependencies are respected, crucial for seamless project continuity.

    \item \textbf{Analyze Time Relationships}: By analyzing time-based dependencies (e.g., FS, SS), this prompt helps identify parallel tasks and branches in dependency graphs, enabling effective time management across activities.

    \item \textbf{Overlapping Disciplines and Inter-Disciplinary Dependencies}: These prompts capture dependencies across overlapping and interconnected disciplines, facilitating resource alignment and identifying areas where interdisciplinary coordination is needed.

    \item \textbf{Area-Based Dependencies}: This prompt encourages the model to examine how dependencies align with specific areas, ensuring location-based planning aligns with the project’s spatial organization.
\end{itemize}

\subsection{Task-Specific Prompts for Data Collection}

For each specific task (Automated Planning (AP), Missing Value Prediction (MVP), Dependency Analysis (DA), and Construction Preference Alignment Direct Preference Optimization (CPA-DPO)), dedicated prompts have been designed to guide the language model in generating relevant outputs. Here’s an outline of each task-specific prompt:

\begin{itemize}
    \item \textbf{Prompt for AP}: This prompt instructs the model to focus on scheduling tasks based on 'Current Start' and 'Current Finish' dates, ensuring that task sequences respect dependencies. By using rules for sequencing and timing, the AP prompt facilitates logical task progression, essential for maintaining project coherence.

    \item \textbf{Prompt for MVP}: This prompt guides the model to predict missing values using both context and generated rules. It emphasizes the identification of critical data points for completion, enhancing data quality and completeness in project tables.

    \item \textbf{Prompt for DA}: Instructing the model to examine dependencies based on 'Predecessor Details' and 'Successor Details,' the DA prompt helps the model identify crucial task interactions. This supports dependency mapping, crucial for understanding the ripple effects of scheduling changes.

    \item \textbf{Context Polishing for CPA-DPO}: This prompt refines the generated output, ensuring it aligns with expert standards. The model adjusts for adherence to preferences, dependencies, and task prioritization, essential for optimized scheduling.
\end{itemize}


Each prompt targets specific construction scheduling needs, aligning outputs with project management best practices and dynamically addressing task complexities.

\subsection{Industry Relevance and Considerations}

The automation of construction scheduling has long been an industry challenge due to the dynamic nature of project constraints, interdependent tasks, and expert-driven decision-making. While traditional methods rely on predefined heuristics and rule-based scheduling, they struggle to adapt to unexpected changes in workforce availability, material delays, or regulatory shifts. Large-scale projects, such as semiconductor fabrication, further complicate scheduling due to high coordination demands across multiple disciplines. Addressing these challenges requires an intelligent, adaptive system capable of learning from past schedules and dynamically updating plans based on new constraints.

A major consideration in adopting LLM-driven solutions for construction is their real-world integration and deployment feasibility. Existing project management software, such as Primavera P6 and BIM-based scheduling tools, is widely used by industry professionals. For AI-driven scheduling to be effective, it must complement these tools rather than replace them. The ability of retrieval-augmented models to incorporate structured industry knowledge and expert-aligned reinforcement learning provides a pathway for seamless integration, allowing construction professionals to leverage AI insights while maintaining human oversight in critical decision-making.

Additionally, concerns about data dependency and scalability must be addressed for broader industry adoption. While proprietary datasets are necessary for high-fidelity scheduling predictions, future research could explore the use of open-source construction datasets or synthetic data generation techniques to improve model robustness across diverse projects. Furthermore, factors such as computational overhead, latency, and cost must be considered in deployment, ensuring that AI-powered scheduling remains practical for real-world applications. By tackling these challenges, LLM-driven scheduling can move from a research prototype to a reliable industry tool that enhances efficiency, reduces project risks, and scales across complex construction environments.

\begin{figure*}[h]
\begin{tcolorbox}[colback=white, colframe=gray!70, colbacktitle=gray!20, coltitle=black, title=Sequential Context (Context 1), width=\textwidth, boxsep=5pt, left=3pt, right=3pt, top=5pt, bottom=5pt]
    \textbf{Activity Sequence and Timing} \newline
    List the sequence of construction activities based on the 'Current Start' and 'Current Finish' dates, ensuring they follow the correct order as indicated by 'Predecessor Details' and 'Successor Details'.
    \tcblower
    \textbf{Calculate Activity Duration} \newline
    Based on the 'Current Start' and 'Current Finish' dates, calculate the duration for each activity and establish the step-by-step timeline for the project.
\end{tcolorbox}
The Sequential Context prompt is designed to capture the linear progression of activities in construction. By focusing on the order and duration of activities, this context prompt aids in generating structured timelines, enabling the model to outline a clear sequence and allocate resources efficiently.
\end{figure*}

\begin{figure*}[h]
\begin{tcolorbox}[colback=white, colframe=gray!70, colbacktitle=gray!20, coltitle=black, title=First-Order Context (Context 2), width=\textwidth, boxsep=5pt, left=3pt, right=3pt, top=5pt, bottom=5pt]
    \textbf{Analyze Time Relationships} \newline
    Analyze the 'Predecessor Details' and 'Successor Details' to determine the time domain relationship between activities. Identify which activities are in parallel and the number of branches in the dependency graph.
    \tcblower
    \textbf{Overlapping Disciplines} \newline
    Identify overlapping disciplines from the 'Discipline' column and infer which dependency graphs can be merged based on this overlap.
    \tcblower
    \textbf{Inter-Disciplinary Dependencies} \newline
    Examine the inter-dependency between different disciplines and create a map of these relationships.
    \tcblower
    \textbf{Area-Based Dependencies} \newline
    Using the 'Area' column, analyze area-based dependencies and how they affect the sequence of construction activities.
\end{tcolorbox}
The First-Order Context prompt focuses on immediate dependencies and relationships between tasks. By analyzing time, disciplinary overlaps, and area-based dependencies, this prompt enables the model to capture critical dependencies that could impact the flow of work and resource allocation across parallel activities.
\end{figure*}

\begin{figure*}[h]
\begin{tcolorbox}[colback=white, colframe=gray!70, colbacktitle=gray!20, coltitle=black, title=Hierarchical Context (Context 3), width=\textwidth, boxsep=5pt, left=3pt, right=3pt, top=5pt, bottom=5pt]
    \textbf{Hierarchical Tree Structure} \newline
    Organize the activities into a hierarchical tree structure based on their WBS and identify any activities that should be sequential but are not currently listed as such.
    \tcblower
    \textbf{Assess Sequence Reconstruction} \newline
    For each activity, determine if the sequence can be recovered from the given data. If not, specify what critical information is missing and suggest how to bridge the identified gaps.
\end{tcolorbox}
The Hierarchical Context prompt helps the model understand hierarchical structures in project planning. By focusing on organizing tasks based on work breakdown structure (WBS), this context prompt aids in identifying gaps in sequencing and structuring project phases logically.
\end{figure*}

\begin{figure*}[h]
\begin{tcolorbox}[colback=white, colframe=gray!70, colbacktitle=gray!20, coltitle=black, title=Automated Planning (AP) Prompts, width=\textwidth, boxsep=5pt, left=3pt, right=3pt, top=5pt, bottom=5pt]
    \textbf{AP - Part 1} \newline
    You are a virtual construction expert collaborating with a larger LLM to automate the construction schedule. Use the 'Current Start' and 'Current Finish' dates in the context to ensure tasks are scheduled based on their dependencies. Explain how the selected rules help guide the automation of task sequencing and timing.
    \tcblower
    \textbf{AP - Part 2} \newline
    Justify why these specific rules and context elements are crucial for automating the schedule. Describe the connection between the context and rules, and provide logical reasoning for why these choices will result in a successful automation process.
\end{tcolorbox}
The AP prompt focuses on scheduling construction activities based on start and finish dates, with an emphasis on the rules that support task sequencing and timing. This prompt aims to ensure coherent automation logic while aligning with project constraints and expert expectations.
\end{figure*}

\begin{figure*}[h]
\begin{tcolorbox}[colback=white, colframe=gray!70, colbacktitle=gray!20, coltitle=black, title=Missing Value Prediction (MVP) Prompts, width=\textwidth, boxsep=5pt, left=3pt, right=3pt, top=5pt, bottom=5pt]
    \textbf{MVP - Part 1} \newline
    Based on the following information, choose the correct values for the missing columns. Return the values as a list, separated by commas, with each value enclosed within [Value] and [/Value] tags. The list should contain exactly three values, corresponding to the columns listed in the same order.
    \tcblower
    \textbf{MVP - Part 2} \newline
    This part provides the row input, static knowledge, and context information that the model will use to identify missing values and fill them accurately.
\end{tcolorbox}
The MVP prompt is essential for accurately predicting missing data in construction tables, using both static knowledge and contextual details. This prompt is designed to help the model make accurate value predictions, enhancing data completeness and reliability.
\end{figure*}

\begin{figure*}[h]
\begin{tcolorbox}[colback=white, colframe=gray!70, colbacktitle=gray!20, coltitle=black, title=Dependency Analysis (DA) Prompts, width=\textwidth, boxsep=5pt, left=3pt, right=3pt, top=5pt, bottom=5pt]
    \textbf{DA - Part 1} \newline
    You are a virtual construction expert collaborating with a larger LLM to analyze dependencies between construction activities. Focus on identifying key dependencies using the 'Predecessor Details' and 'Successor Details' in the context. Explain how and why the selected rules are relevant for understanding the dependencies between activities.
    \tcblower
    \textbf{DA - Part 2} \newline
    Connect these rules to specific parts of the context. Ensure that the relationship between the context and rules is clearly articulated, showing logical reasoning behind the choices made for this analysis.
\end{tcolorbox}
The DA prompt guides the model in identifying and explaining dependencies between construction activities, with emphasis on critical tasks and their interactions. This prompt supports dependency mapping, which is crucial for project planning and risk management.
\end{figure*}


\begin{figure*}[h]
\begin{tcolorbox}[colback=white, colframe=gray!70, colbacktitle=gray!20, coltitle=black, title=Context Polishing for CPA-DPO Prompts, width=\textwidth, boxsep=5pt, left=3pt, right=3pt, top=5pt, bottom=5pt]
    \textbf{Context Polishing for CPA-DPO - Part 1} \newline
    As a virtual construction scheduling expert, refine the following output to ensure it aligns with expert expectations. Your role involves guiding a larger LLM by providing clear context, expert rules, and structured instructions for three primary tasks:
    \begin{itemize}
        \item \textbf{Missing Value Prediction:} Select and explain relevant context elements crucial for filling in missing values. Use expert rules to guide predictions and clarify their connection to the context.
        \item \textbf{Dependency Analysis:} Analyze and explain activity dependencies using 'Predecessor Details' and 'Successor Details.' Highlight how the rules inform these relationships.
        \item \textbf{Schedule Automation:} Automate task scheduling using 'Current Start' and 'Current Finish' dates, prioritizing based on criticality and dependencies. Apply rules to ensure task order and dependencies are respected.
    \end{itemize}
    \tcblower
    \textbf{Context Polishing for CPA-DPO - Part 2} \newline
    The output should provide coherent and contextually relevant responses to scheduling needs, integrating expert rules and project-specific knowledge seamlessly. Emphasize adherence to preferences and explain any dependencies or task prioritizations that support an optimized construction schedule.
\end{tcolorbox}
The Context Polishing prompt ensures that responses align with expert preferences, providing clear, structured guidance for missing value prediction, dependency analysis, and schedule automation. It supports the Direct Preference Optimization (DPO) process by enhancing the alignment of generated content with real-world project standards and expectations.
\end{figure*}




\end{document}